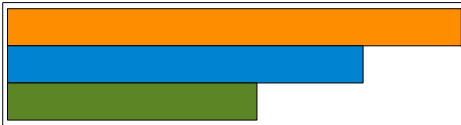

# Networks Routing Optimization

## Using Swarm Intelligence

By Mohamed Hassan

## 1- Introduction

The aim of this paper is to highlight and explore a traditional problem, which is the minimum spanning tree, and finding the shortest-path in network routing, by using Swarm Intelligence. In other words this work to be considered as an investigation topic with combination between operations research, discrete mathematics, and evolutionary computing aiming to solve one of networking problems.

Minimum spanning tree and path-finding (shortest one) in a graph mathematically have been around for a while, already there are many contribution, but a few searching to find a fast, new, and easy way especially to solve the multiple destination routing (MDR) problems. So we will try to explore this topic with an integration between traditional algorithm and particle swarm optimization (PSO) algorithm. The PSO algorithm is based on the sociological behavior associated with bird flocking or fishing school. The algorithm makes use of cognitive and social information among the individuals (particles) to find an optimal solution to an optimization problem.

PSO is an important swarm intelligent algorithm with fast convergence speed and easy implementation. We start to discover what is the evolutionary computing, as well as what is the genetic algorithm; the concept of PSO (Particle Swarm Optimization) as a part of new methods in computation (Evolutionary Computing), in addition we will describe and analyze the spanning tree and network route optimization, to get the minimal spanning tree, and finally finding the shortest route using swarm intelligence represented in Particle Swarm Optimization (PSO).

Let's now dig into the methods and explore the differences and advantages of each.

A network consists of a set of nodes connected together. Here we study the network in generalized context. We will give a simple notation for our network which is (V, E), where V is the number of Vertices (nodes) and E is the set of Edges (Links/Cables). Example figure 2.1

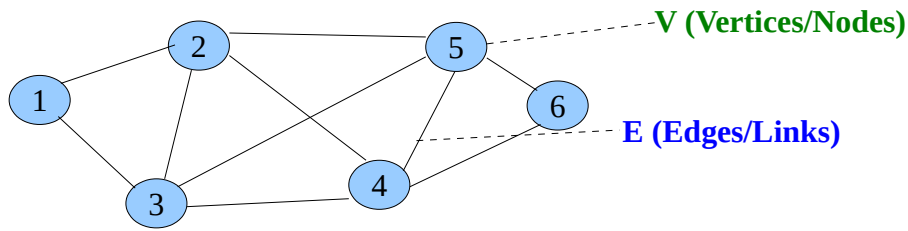

Figure 2.1

**V** = {1,2,3,4,5}, **E**={(1,2),(1,3),(2,3),(2,4),(2,5),(3,4),(3,5),(4,5)}.

Each link (path) is assigned a capacity which may be flow, cost, distance, etc ... We define a path as a sequence of distinct links that join two nodes of the flow irrespective of the direction of flow in each link. If the path connects a node and itself we consider the path to be cyclic, as in figure 2.1 links (3,4), (4,5) and (5,3) form a closed loop (cycle). In particular, a directed cycle is a loop in which all links are directed in the same direction. Nodes which are connected without loop are a-cyclic. In other word a connected network means that any pair of distinct nodes can be connected by at least one path. A tree is connected network without loops. [1]

## 2. A brief History About Evolutionary Inspired Computing

Biological behavior has inspired researchers to adapt the natures in computing, their work is not mature enough, but it is pushing technology to its most interesting limits.

(A) Genetic Programming (GP):
Is an implementation of evolutionary programming, where the problem-solving domain is modeled on computer and the algorithm attempts to find a solution by the process of simulated evolution, employing the biological theory of genetics and the Darwinian principle of survival of the fittest. GP is distinct from other techniques because of its tree representation and manipulation of all solutions.

(B) Particle Swarm Optimization (PSO):
Similarly to Genetic Algorithms, PSO is a population-based technique, inspired by the social behavior of individuals (or particles) inside swarms in nature (for example, flocks of birds or schools of fish). PSO first proposed by Kennedy and Eberhart (1995). In addition PSO is a collection of particles or (agents) swarm through an N-dimensional space. The rules for how the particles move through the space are based on simple natural flocking rules that cause the particles to orbit around the best-found solution in the hope of finding a better one.[2]
So particles will usually looking forward for the best (optimum) solution. PSO is a simple algorithm (method), as well as fast, effective, and can be used in many types of optimization problems.

## 3. Network optimization (spanning tree and shortest path) – and its Algorithms

A network consists of a set of nodes (vertices) linked by arcs (edges) and a network optimization models actually are special types of linear programming problems.
In the next few papers we only scratch the surface of the current network route overview. However, we proposed some basic ideas of how to solve them using PSO (without delving into issues of data structures that are so vital to successful large-scale implementations).
The minimum spanning tree problem bears some similarities to the shortest-path problem.

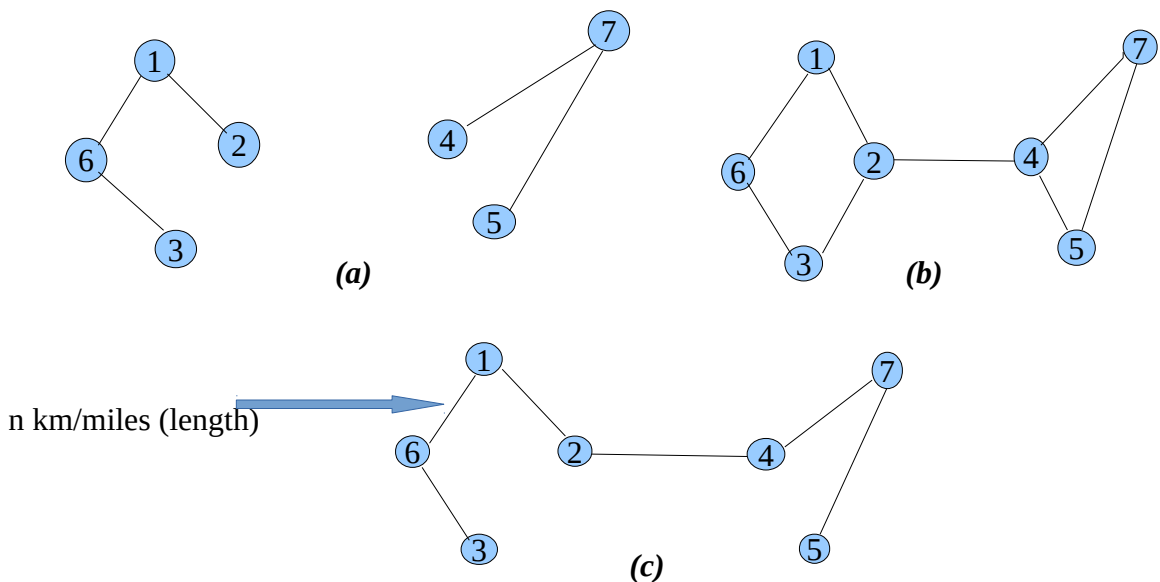

*Figure 3.1 Shows the spanning tree concept*
*(a)* Not a spanning tree, *(b)* Not a spanning tree, *(c)* A spanning tree

A network with *n* nodes requires only *(n – 1)* links to provide a path between each pair of nodes. *Figure 3.1* illustrates this concept of a spanning tree for the a sample network diagram. Thus, *Fig. 3.1 (a)* is not a spanning tree because nodes 6, 1, 2, and 3 are not connected with nodes 4, 5, and 7. It needs another link to make this connection. This network actually consists of two trees, one for each of these two sets of nodes. The links in *Fig. 3.1 (b)* do span (the network is connected), but it is not a tree because there are two cycles (6–1–2–3–6 and 4–7–5–4). It has too many links. Because this network problem sample has n = 7 nodes, and we already indicates that the network must have exactly *n – 1 = 6* links, with no cycles, to qualify as a spanning tree. This condition is achieved in *Fig. 3.1 (c)*, so this network is a feasible solution (with a value of n km/miles for the total length of the links) for the minimum spanning tree problem. In short a tree is a connected nodes without cycles.[3]

## 3.1 Algorithm for the Minimum Spanning Tree & Path-finding

<u>Objective of $n^{th}$ iteration:</u> Find the $n^{th}$ nearest node to the origin (to be repeated for *n = 1, 2, . . .* until the $n^{th}$ nearest node is the destination.
<u>Input for $n^{th}$ iteration:</u> *(n – 1)* nearest nodes to the origin (solved for at the previous iterations), including their shortest path and distance from the origin. (These nodes, plus the origin, will be called solved nodes; the others are unsolved nodes).
<u>Candidates for $n^{th}$ nearest node:</u> Each solved node that is directly connected by a link to one or more unsolved nodes provides one candidate the unsolved node with the shortest connecting link. (Ties provide additional candidates).
<u>Calculation of $n^{th}$ nearest node:</u> For each such solved node and its candidate, add the distance between them and the distance of the shortest path from the origin to this solved node. The candidate with the smallest such total distance is the $n^{th}$ nearest node (ties provide additional solved nodes), and its shortest path is the one generating this distance.

## 3.2 Traditional Algorithms for Minimum Spanning Tree & Path-finding

## 3.2.1 (Prim's Algorithm)

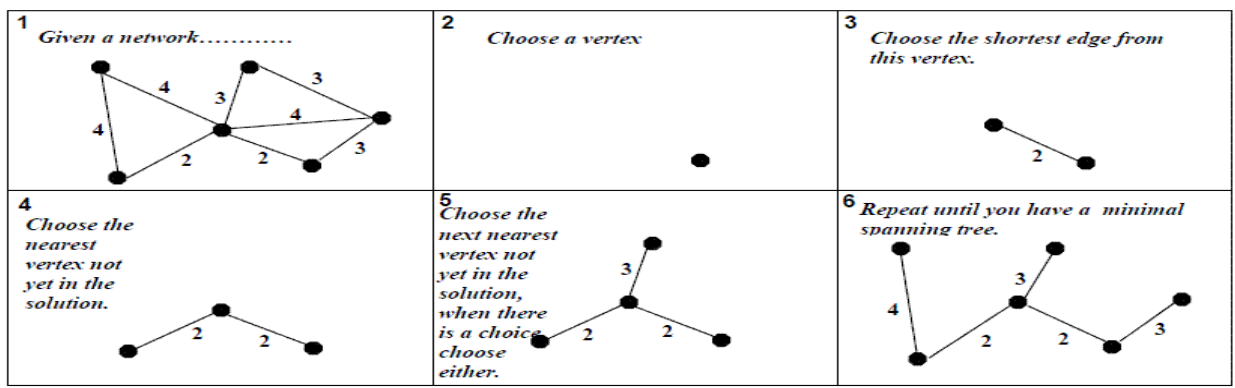

Source [4]

### *3.2.2 (Kruskal's Algorithm)*

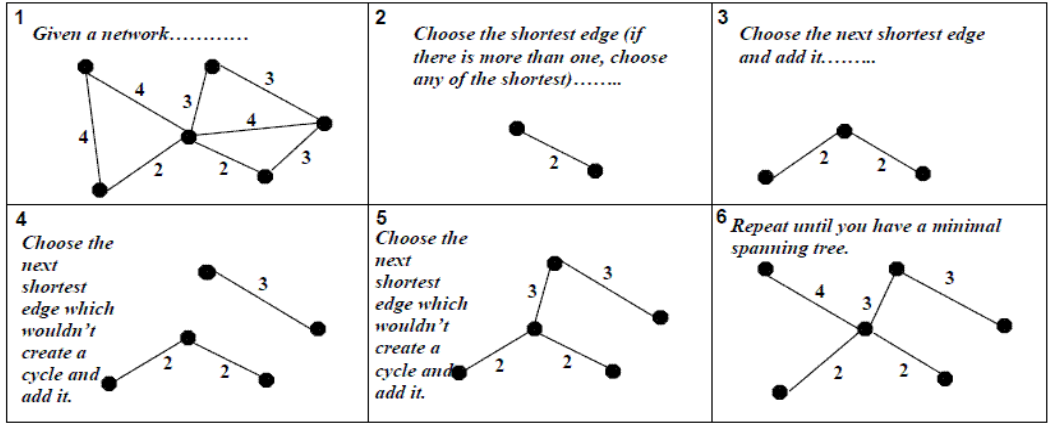

Source [5]

### *3.3 Why minimum spanning trees?*

The standard application is to a problem like phone network design. You have a business with several offices; you want to lease phone lines to connect them up with each other; and the phone company charges different amounts of money to connect different pairs of cities. You want a set of lines that connects all your offices with a minimum total cost. It should be a spanning tree, since if a network isn't a tree you can always remove some edges and save money.[6]

- They can be computed quickly and easily.
- They provide a way to identify clusters in sets of points. Deleting the long edges from a minimum spanning tree leaves.
- They can be used to design of telecommunication networks (fiver-optic networks, computer networks, telephone networks, television networks, etc..)
- As an educational tool, minimum spanning tree algorithms provide graphic evidence that greedy algorithms can give provably optimal solutions.

### *3.4 Why using PSO?!*

While traditional method to solve such as above problems performs a limited exploration of search space and result in inferior solutions, some other traditional method are general purpose optimization that usually involve a large set of parameters that needs to be fine-tuned, thus PSO (particle swarm optimization) is an extremely simple algorithm that seems to be effective for optimizing a wide range of functions. We view it as a mid-level form a natural life or biologically derived algorithm, occupying the space in nature between evolutionary search, which requires eons, and neural processing, which occurs on the order of milliseconds. Social optimization occurs in the time frame of ordinary experience, in fact, it is ordinary experience. In addition to its ties with natural life, particle swarm optimization has obvious ties with evolutionary computation. Conceptually, it seems to lie somewhere between genetic algorithms and evolutionary programming. It is highly dependent on stochastic processes, like evolutionary programming.

We summarize the advantage of using PSO as follow:

- Insensitive to scaling of design variables
- Simple implementation
- Easily parallelized for concurrent processing
- Derivative free
- Very few algorithm parameters
- Very efficient global search algorithm

[7]

## 4. Using Paticle Swarm Optimization (PSO) for Network Optimization

In addition of the above particle swarms have a nutral motivation. Two specific inspiration are influnce of PSO fish school, and bird flocks. Each involoves collection (groups) of elements (particles) that move together synchronously, diverge from one to another, and then re-group. Ultimately we just say particle swarms as a means to simulate social behavior by using computer-based model, hence we will refer to PSO in the next few papers as algorithm (PSO Algoritm). [8]

## 4.1 PSO Algorithm

The PSO is a population-based algorithm in which individual particles work together to solve given problem. Now we will walk through the basic PSO algorithm and take a look at each of its elements. The most interesting thing about PSO algorithm is its pure simplicity. The PSO can be implemented in very little code using only primitive mathematical operators, making it ideal for limited-memory and computationally constrained environments.[9]

## 4.1.1 Basic PSO Flow

The basic flow for the PSO algorithm:

1- As we begin by random particle(s) initialization (we call it population) within N-dimensional space.
2- Because it random population, so each particle will have a random location and velocity for each population dimension (space).
3- Then we will evaluate each particle's fitness for any given problem (objective function), if fitness is better than the personal particle best fitness (pbest), we will save the location vector for that particle as pbest. If the particle's fitness is better than the global best fitness (gbest), we save this particle's location vector as gbest.
4- Finally, we update the particle's velocity and position and look at the next particle in the population. If the global best fitness meets the exit criteria (satisfactory criteria – best fitness of all particle's), we end the run (loop) and provide the location vector as the solution to the given problem, each element of the solution vector represents the independent variable for given problem.[10]

## *4.1.2 How it works?*

Each particle has a position (x, y) in the search space and a velocity (vx, vy) at which it is moving through that space (N-dimensional). Particles have a certain amount of inertia, which keeps them moving in the same direction they were moving previously.
They also have acceleration (change in velocity), which depends on two main things.

1) Each particle is attracted toward the best location that it has personally found (personal best - *pbest*) previously in its history.

2) Each particle is attracted toward the best location that (any) particle has ever found (global best - *gbest*) in the search space.

The strength with which the particles are pulled in each of these directions is dependent on the parameters (pbest) and (gbest). As particles move farther away from these "best" locations, the force of attraction grows stronger, there is also a random factor about how much the particle is pulled toward each of these locations, the particle swarm is trying to optimize a function that is determined by the values in the discrete grid of cells, we will describe that in more details in the next sections. [11]

## *4.2 THINGS TO NOTICE*

You will often see particles traveling in paths that are roughly elliptical. Sometimes the swarm quickly finds the perfect solution, and other times it becomes stuck in the wrong area of the search space, and looks like it may never find the perfect solution. One variation of the PSO algorithm uses a repulsive force between particles to help keep them spread out in the space, and less likely to all gravitate to a sub-optimal value.

## *4.3 The Problem (Example)*

Consider an ISP network. At any given time, a packet sent by one computer typically traverses multiple routers before reaching its destination, and it may take a certain amount of time to traverse each path (due to congestion effects, switching delays, and so on), making communication performance depend on the flow of traffic. We consider the role of optimization in controlling the routing of traffic through the Internet. Assuming the following (*Figure 4.1*) network type, and we need to find the shortest path. In such a problem, you have a network with link weight on the edges and two special nodes: a start node (White) and a finish node (Green).

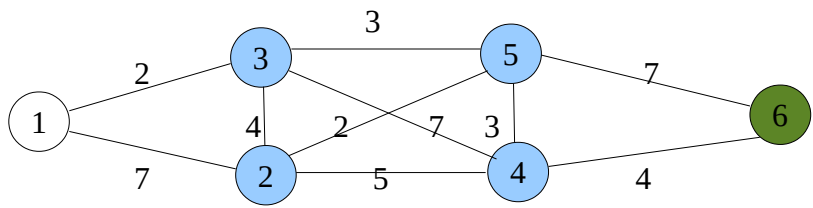

*Figure 4.1*

PSO is used to either maximize or minimize and objective function by updating the velocity (cost, weight, link) in every iteration. We will map (consider) all paths from Start (Origin) to Finish (Destination) as particles.

*Things to be considered :*

A) <u>Parameters such as</u>: Every particle is given a random *position* and *velocity* (Initialization).
B) The *pBest* value is calculated depending on the value of fitness of each particle in every iteration.
C) The *fitness* value depends upon the *weight* associated with the particles.

$$Fitness = \frac{pBest_{StartWithShortest}}{\sum nP} \quad >> where\ n\ (number)\ and\ P\ (Particle)$$

*pBest or pbest will be considered our <u>priority</u>.*

## *4.4 PSO Simple Algorithm to Optimize Network Routing*

1- Calculate *pbest* and position of corresponding particle.
2- for(i=0; i<no_of_iteration; i++);
3- for(x=0; x<particle_count; x++);
4- Change weight of each link along the path of each particle randomly.
5- Calculate fitness and find *pbest* of $n^{th}$ iteration.
6- Find best *pbest* or *gbest* with the corresponding particle, which gives the shortest-path in terms of the weight associated with the links on the path (or particle). Figure 4.1

## *4.5.1 illustration of how it works*

<u>Note:</u> pbest (priority) is not the weight Figure 4.1.

| Node | 1 | 2 | 3 | 4 | 5 | 6 |
|---|---|---|---|---|---|---|
| Pbest (priority) | 2 | 6 | 4 | 9 | 5 | 7 |

$i^0 \qquad\qquad\qquad\qquad i=0, V_p^k=(1)$

| Node | 1 | 2 | 3 | 4 | 5 | 6 |
|---|---|---|---|---|---|---|
| Pbest (priority) | ● | 6 | ● | 9 | 5 | 7 |

$i^1 \qquad\qquad\qquad\qquad i=1, V_p^k=(1,3)$

| Node | 1 | 2 | 3 | 4 | 5 | 6 |
|---|---|---|---|---|---|---|
| Pbest (priority) | ● | ● | ● | 9 | 5 | 7 |

$i^2 \qquad\qquad\qquad\qquad i=2, V_p^k=(1,3,2)$

| Node | 1 | 2 | 3 | 4 | 5 | 6 |
|---|---|---|---|---|---|---|
| Pbest (priority) | ● | ● | ● | 9 | ● | 7 |

$i^3 \qquad\qquad\qquad\qquad i=3, V_p^k=(1,3,2,5)$

| Node | 1 | 2 | 3 | 4 | 5 | 6 |
|---|---|---|---|---|---|---|
| Pbest (priority) | ● | ● | ● | ● | ● | 7 |

$i^4 \qquad i=4, V_p^k=(1,3,2,5,4)$

| Node | 1 | 2 | 3 | 4 | 5 | 6 |
|---|---|---|---|---|---|---|
| Pbest (priority) | ● | ● | ● | ● | ● | ● |

$i^5 \qquad i=5, V_p^k=(1,3,2,5,4,6)$

*Particle priority vector updating to construct the path {1, 3, 2, 5, 4, 6}.*

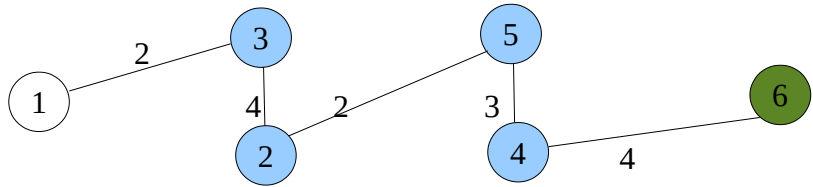

*Figure 4.2*

This example corresponds to the priority (particle position) vector to create a valid path from the same position vector and the corresponding final path is {1, 3, 2, 5, 4, 6}, where forces the algorithm to choose a valid node (although it has the highest priority) thereby avoiding an invalid path creation. This operation reduces the number of invalid paths (thereby, the computation time).[12]

## *5. Conclusion*

The purpose of this paper is to investigate and highlight a specific area in computer science with combination of the intelligence behavior of biological swarm and operations research trying to optimize network routing focusing on how to minimal spanning tree and solve the shortest path problem using Particle Swarm Optimization (PSO) techniques. PSO algorithm is simple and efficient method for optimization and discontinuous multidimensional problems. PSO in general can be used interchangeably with the genetic algorithm, and in some cases, can find solutions faster with less computational overhead.

An endeavor was made to reduce the probability of invalid loop/backward path, as will as find the valid and near optimal path with PSO based search techniques with less iteration process. As each particle chooses explicitly its location based on its personal best (pbest) from the set of particle's global best (gbest) in the entire swarm, accordingly the near shortest-path should be chosen in less computation time and of-course with less code. Since this paper focus on highlighting of how pso can be used to solve optimization problem, so it should be further work to include codes/application and fine tuning to solve more network problem such as network flow.